\documentclass[letterpaper]{article} 
\usepackage{aaai2026}  
\usepackage{times}  
\usepackage{helvet}  
\usepackage{courier}  
\usepackage[hyphens]{url}  
\usepackage{graphicx} 
\usepackage{natbib}  
\usepackage{caption} 
\usepackage{amsmath}
\usepackage{amssymb}
\usepackage{gensymb}
\usepackage{booktabs}
\usepackage{tabularx}
\usepackage[position=top]{subfig}
\usepackage{algorithm}
\usepackage{algpseudocode}

\usepackage{hyphenat}
\usepackage[shortcuts]{extdash}

\usepackage{amsthm}

\newtheorem{theorem}{Theorem}
\newtheorem{lemma}{Lemma}
\newtheorem{definition}{Definition}
\newtheorem{question}{Research Question}
\newtheorem{proposition}{Proposition}

\title{Self-Supervised Inductive Logic Programming}

\author {
	Stassa Patsantzis 
}
\affiliations{
   German University of Digital Sciences \\
   stassa.patsantzis@german-uds.de 
}

\begin{document}

\maketitle

\begin{abstract}
Inductive Logic Programming (ILP) approaches like Meta \-/ Interpretive Learning
(MIL) can learn, from few examples, recursive logic programs with invented
predicates that generalise well to unseen instances. This ability relies on a
background theory and negative examples, both carefully selected with expert
knowledge of a learning problem and its solutions. But what if such a
problem-specific background theory or negative examples are not available? We
formalise this question as a new setting for Self-Supervised ILP and present a
new MIL algorithm that learns in the new setting from some positive labelled,
and zero or more unlabelled examples, and automatically generates, and labels,
new positive and negative examples during learning. We implement this algorithm
in Prolog in a new MIL system, called Poker. We compare Poker to
state-of-the-art MIL system Louise on experiments learning grammars for
Context-Free and L-System languages from labelled, positive example strings, no
negative examples, and just the terminal vocabulary of a language, seen in
examples, as a first-order background theory. We introduce a new approach for
the principled selection of a second-order background theory as a Second Order
Definite Normal Form (SONF), sufficiently general to learn all programs in a
class, thus removing the need for a backgound theory tailored to a learning
task. We find that Poker's performance improves with increasing numbers of
automatically generated examples while Louise, bereft of negative examples,
over-generalises.
\end{abstract}

%
\begin{links}
	\link{Code}{https://github.com/stassa/aaai_26_experiments/}
	\link{Appendices}{https://arxiv.org/abs/2507.16405}
\end{links}

\section{Introduction}
\label{Introduction}

\begin{table*}[t]
	\small
	\centering
	\setlength{\tabcolsep}{1.6mm}
	\begin{tabularx}{0.99\textwidth}{cccccccc}
		\multicolumn{8}{c}{ \textbf{Self-supervised ILP with Poker} } \\
		\toprule
		\textbf{Labelled examples} $\mathbf{E^+{:}1^n0^n (n > 0)}$ & \multicolumn{7}{c}{\textbf{Unlabelled examples} $\mathbf{E^?{:}1^n0^m (n \geq m \geq 0)}$ (21 of 100 for $n \in [0,18]$) } \\
		$s(1^1,0^1).$ & $s(1^0,0^0).$ & $s(1^3,0^0).$ & $s(1^6,0^0).$ & $s(1^9,0^9).$ & $s(1^7,0^7).$ & $s(1^6,0^6).$ & $s(1^5,0^3).$\\
		$s(1^2,0^2).$ & $s(1^1,0^0).$ & $s(1^4,0^0).$ & $s(1^7,0^0).$ & $s(1^8,0^8).$ & $s(1^6,0^1).$ & $s(1^5,0^1).$ & $s(1^5,0^5).$\\
		$s(1^3,0^3).$ & $s(1^2,0^0).$ & $s(1^5,0^0).$ & $s(1^8,0^0).$ & $s(1^7,0^1).$ & $s(1^6,0^2).$ & $s(1^5,0^2).$ & $s(1^4,0^1).$\\
	\end{tabularx}
	\begin{tabular}{rlll}
		\multicolumn{4}{c}{ $\mathbf{B:}$ \textbf{First-order background theory} } \\
		& $one([1|X], X).$      & $zero([0|X], X).$     & $empty(X, X).$ \\
	\end{tabular}
	\begin{tabularx}{\textwidth}{lX}
		\multicolumn{2}{c}{ $\mathbf{\mathcal{M}:}$ \textbf{Second-Order Background Theory} } \\
		\textbf{(Identity)} $P(x,y) \leftarrow Q(x,y):$ 	& $target(P) $ $\wedge \; background(Q) \vee empty(Q) $ \\
		\textbf{(Chain)} $P(x,y) \leftarrow Q(x,z), R(z,y):$	& $P \neq Q $ $\wedge \; (target(P) \vee invented(P)) $ $\wedge \; not(target(Q)) $ $\wedge \; not(empty(Q,R)) $ \\ 
									& $\wedge \; invented(P,Q) \rightarrow P \neq Q $\\
	\midrule
	\end{tabularx}
	\begin{tabular}{llll}
		\multicolumn{4}{c}{ \textbf{Learned Hypothesis} } \\
		With invented predicates ($s_1/2$): & $s(A,B) \leftarrow one(A,C), zero(C,B).$ & $s(A,B) \leftarrow s_1(A,C), zero(C,B).$ \\ & & $s_1(A,B) \leftarrow one(A,C), s(C,B)).$ \\
		Unfolded to remove invented predicates: & $s(A,B) \leftarrow one(A,C), zero(C,B).$ & $s(A,B) \leftarrow one(A,C), s(C,D), zero(D,B).$ & \\
	\end{tabular}
	\begin{tabularx}{\textwidth}{lll|lllllll}
		\midrule
		\multicolumn{10}{c}{{\hspace{5em}} \textbf{Labelling} (includes both automatically generated and user-given unlabelled examples) } \\
		\multicolumn{3}{c}{ \textbf{Labelled Positive} } & \multicolumn{7}{c}{ \textbf{Labelled Negative} (21 of 175) } \\
		$s(1^1,0^1).$ & $s(1^4,0^4).$ & $s(1^7,0^7).$ & $s(1^0,0^0).$ & $s(1^2,0^1).$ & $s(1^3,0^2).$ & $s(1^4,0^2).$ & $s(1^5,0^1).$ & $s(1^6,0^0).$ & $s(1^7,0^0).$ \\
		$s(1^2,0^2).$ & $s(1^5,0^5).$ & $s(1^8,0^8).$ & $s(1^1,0^0).$ & $s(1^3,0^0).$ & $s(1^4,0^0).$ & $s(1^4,0^3).$ & $s(1^5,0^2).$ & $s(1^6,0^1).$ & $s(1^7,0^1).$ \\
		$s(1^3,0^3).$ & $s(1^6,0^6).$ & $s(1^9,0^9).$ & $s(1^2,0^0).$ & $s(1^3,0^1).$ & $s(1^4,0^1).$ & $s(1^5,0^0).$ & $s(1^5,0^3).$ & $s(1^6,0^2).$ & $s(1^8,0^0).$ \\
		\bottomrule
	\end{tabularx}
	\caption{Poker inputs and outputs. Example strings pretty-printed from DCG
	notation (e.g. $s([1,1,0,0],[])
	\rightarrow s(1^2,0^2)$).}
	\label{tab:poker_initial_example}
\end{table*}

In the Standard Setting \cite{Nienhuys-Cheng1997} for Inductive Logic
Programming \cite{Muggleton1991} (ILP) a background theory $B$ and positive and
negative examples $E^+, E^-$ are used as training data to learn a \emph{correct}
hypothesis $H$ that accepts, in conjunction with $B$, all examples in $E^+$, and
no examples in $E^-$. Typically, $E^+, E^-$ are selected, and $B$ programmed,
manually by the user according to their background knowledge of the learning
target (i.e. the logic program to be learned). $B$ in particular is tailored to
each learning target \cite{Flener1999}, and $E^-$ hand-picked to avoid
over-generalisation. Given the right $B,E^+$ and $E^-$, recent ILP systems can
learn correct hypotheses with recursive clauses and invented predicates.

The ability to include background knowledge in training data is a desirable
characteristic of ILP systems, but the practice of manually creating a
target-specific background theory and selecting negative examples is a constant
burden that limits the real-world application of ILP approaches.

In this paper we present a new way to alleviate the aforementioned burden on the
user with a new algorithm for Self-Supervised ILP, more specifically,
Self-Supervised Meta-Interpretive Learning\cite{Muggleton2014} (MIL). The new
algorithm, implemented in a new MIL system called Poker\footnote{Named after,
not the game, but \emph{Wittgenstein's Poker}; a friendly dig at Popper
\cite{Cropper2021}.}, is ``self-supervised" in the sense that it learns from
both labelled positive and unlabelled examples and it automatically generates
new positive and negative examples during learning. Because it generates new
negative examples, Poker can learn from a background theory that is not tailored
to the learning target without over-generalising. Poker returns not only a
hypothesis, but also a \emph{labelling} of unlabelled, and automatically
generated, examples. 

We illustrate the use of Poker in Table \ref{tab:poker_initial_example} where
Poker is given 3 positive, labelled examples, $E^+$, of the $\{1^n0^n: n > 0\}$
($1^n0^n$) Context-Free Language (CFL), and 30 unlabelled examples, $E^?$ of the
$\{1^n0^m: n \geq m \geq 0\}$ ($1^n0^m$) CFL which includes $1^n0^n$ as a
subset. Examples are atoms in Definite Clause Grammars (DCG) notation
\cite{Pereira1987} representing strings of $1^n0^n$ and $1^n0^m$. The
first-order background theory, $B$, consists of just the terminal vocabulary of
both languages, $\{1,0,\epsilon\}$, defined as a set of DCG pre-terminals.
$1^n0^n$ and $1^n0^m$ are more often denoted as $a^nb^n$ and $a^nb^m$
respectively, but replacing $a,b$ with $1,0$ makes it possible for $B$ to
express any grammar with two terminal symbols suitably mapped to $1$ and $0$.
$B$ can also be constructed \emph{automatically} from $E^+,E^?$. A second-order
background theory, $\mathcal{M}$, includes two \emph{metarules}, \emph{Chain}
and \emph{Identity}, Second-Order definite clauses with a set of constraints
encoding a \emph{Second Order Definite Normal Form} (SONF)\footnote{SONFs are
formalised in the Framework Section.}. The background theory $B \cup
\mathcal{M}$ is thus sufficiently general to express, as a DCG, a Context-Free
Grammar (CFG) of any \emph{bit-string} CFL or Regular language, i.e. one with a
vocabulary of at most two characters and $\epsilon$.

The maximal generality of $B \cup \mathcal{M}$ in Table
\ref{tab:poker_initial_example} achieves two purposes. On the one hand it
guarantees that $B \cup \mathcal{M}$ is general enough to learn the target
grammar, i.e. a CFG of $1^n0^n$. On the other hand, $B \cup \mathcal{M}$ is no
longer target specific: instead of being tailored to one learning target,
$1^n0^n$, it can be reused to learn any bit-string CFG. At the same time, the
generality of $B \cup \mathcal{M}$ introduces a problem: in the absence of
negative examples, it is impossible to distinguish $1^n0^n$ from $1^n0^m$ (the
language of $E^?$).  Indeed, without negative examples it is impossible to
distinguish any bit-string language from $\{0,1\}^*$ the maximally general
language of all bit-strings. In order to learn $1^n0^n$ without
over-generalising, therefore, negative examples are necessary. Poker can
generate negative examples automatically, thus avoiding over-generalisation.

It is also possible to avoid over-generalisation by learning a hypothesis that
only accepts $E^+$, i.e. over-fitting $E^+$. Poker returns a labelling of $E^?$
which, in Table \ref{tab:poker_initial_example}, include $1^n0^n$ strings that
are not in $E^+$. In order to correctly label examples in $E^?$ Poker must
therefore learn a hypothesis that generalises at least to the $1^n0^n$ strings
in $E^?$.

In summary, Poker's ability to automatically generate negative examples makes it
possible to use a maximally general background theory that is no longer tailored
to a single learning target. This ability frees the user from having to manually
select a background theory and negative examples for each new learning target.

\subsection{Self-Supervised ILP by detection of contradictions}
\label{Self-Supervised ILP by detection of contradictions}

The intuition behind Poker's algorithm is that, if two atoms (in the First Order
Logic sense) $e_1$ and $e_2$ are accepted by the same hypothesis $H$ (a logic
program), i.e. $H \models \{e_1, e_2\}$, then to assume that $e_1$ is a positive
and $e_2$ a negative example of $H$ is a contradiction (in the informal sense).
Such a contradiction can be detected in the following manner. Suppose $T =
\{H_1, ..., H_m\}$ is a set of hypotheses and $T \models \{e_1, e_2\}$. And
suppose that we remove from $T$ each $H_i$, where $H_i \models e_2$ leaving
behind the set $T'$ such that $T' \not \models e_2$. There are now two
possibilities: either $T' \models e_1$, in which case there is no contradiction;
or $T' \not \models e_1$, in which case there is a contradiction: $e_1$ and
$e_2$ are both accepted by some subset of $T$, now missing from $T'$; therefore,
$e_2$ is a positive, not a negative, example of the subset of $T$ that accepts
$e_1$. 

Accordingly, Poker begins by constructing a set $T$ of initial hypotheses that
accept the labelled examples $E^+$. Poker can generate new unlabelled examples,
added to $E^?$, by executing $T$ as a generator. Poker then assumes that each
unlabelled example $e^? \in E^?$ is negative and removes from $T$ each
hypothesis $H$ that accepts $e^?$ in conjunction with $B$. If the remaining $T$
now rejects any examples in $E^+$, $e^?$ is re-labelled as positive and moved to
$E^+$. The labelling process thus iteratively \emph{specialises} $T$ until it is
consistent with $E^+$. The labelling process is not without error but its
accuracy increases monotonically with the cardinality of $E^?$.

\subsection{Contributions}

We make the following contributions:


\begin{itemize}
	\item A new setting for Self-Supervised ILP.
	\item A new MIL algorithm for Self-Supervised ILP, and a new MIL system,
		Poker, implementing the new algorithm.
	\item A definition of Second-Order Definite Normal Forms (SONFs), a new
		kind of second-order background theory sufficiently general to
		learn all programs in a class.
	\item Two SONFs for CFGs and L-System grammars in DCG notation.
	\item A proof that Poker's accuracy increases monotonically with the
		number of unlabelled examples.
	\item Experiments investigating the effect of automatically generated
		examples on Poker's learning performance.
\end{itemize}

\section{Related Work} 
\label{Related Work}

\subsection{Self-Supervised Learning}
\label{Self-Supervised Learning}

Self-Supervised Learning is introduced in \cite{Raina2007}, albeit under the
rubric of ``Self-Taught Learning'', where few, labelled, positive examples are
used along with a larger number of unlabelled examples to learn discriminative
features for ``downstream'' supervised learning tasks, particularly image
classification. Later approaches expect only unlabelled examples but generate
negative examples automatically by means of ``pretext'' tasks \cite{Gui2023ASO},
such as image translations or rotations, as in Contrastive Learning
\cite{Hu2024}. Choosing pretext tasks requires domain knowledge and generated
examples are not used in downstream supervised learning tasks. Poker instead
automatically generates both positive and negative examples and labels them
consistently with labelled examples without the need of pretext tasks.
Additionally Poker does not learn features for downstream learning tasks but
directly uses its automatically generated examples to learn arbitrary logic
programs. Logic programs learned by Poker are not classifiers but instead can be
executed as either discriminators or generators as we demonstrate in the
Experiments Section.


\subsection{Self-Supervised Learning in ILP}
\label{Self-Supervised Learning in ILP}

We are not aware of earlier work on self-supervised ILP. The problem of ILP's
over-reliance on manually defined background theories and negative examples has
been tackled before.

Learning from a single example, a.k.a. one-shot learning, is comparable to
self-supervised learning, in that only a single example is given that is assumed
to be positive and there are no negative examples. Numerous works in the MIL
literature demonstrate learning from positive-only, or single examples, e.g.
\cite{Lin2014,DaiWangZhou2018}. However, those systems rely on a
problem-specific background theory selected manually and ad-hoc to avoid
over-generalisation. In this paper instead we introduce the use of Second Order
Definite Normal Forms as a new, principled way to manually select a second-order
background theory for MIL.

DeepLog \cite{Muggleton2023} is a recent MIL system that explicitly learns from
single examples, and a standard library of low-level primitives that apply to
entire classes of problems, instead of a problem-specific background theory.
DeePlog is limited to second-order background theories in dyadic logic (with
literals of arity 2). Instead Poker's SONFs are not limited by arity, or other
syntactic property.

Popper \cite{Cropper2021} is a system of the recent Learning from Failures
setting for ILP that aims to learn logic programs without recourse to a
user-provided higher-order background theory, such as required by Poker and
other MIL systems. However, Popper still needs a problem-specific first-order
background theory, extra-logical syntactic bias in the form of mode
declarations, and negative examples. The need for negative examples is a
limitation addressed in \cite{Yang2025Ext}\footnote{The authors consider the
need for negative examples a limitation of ``classic'' or ``standard'' ILP by
which they seem to refer specifically to Popper; that is an
over-generalisation.}. The approach in \cite{Yang2025Ext} only applies to
recursive data structures like lists or heaps. Instead Poker's algorithm learns
arbitrary logic programs with recursion and invented predicates.

Meta$_{Abd}$ \cite{DaiWangZhou2021} is a neuro-symbolic MIL system that trains a
deep neural net in tandem with a MIL system modified for abductive learning, on
unlabelled, sub-symbolic examples. Meta$_{Abd}$ relies on a manually defined,
problem-specific first-order background theory in the form of definitions of
adbucible predicates. Instead Poker does not require a problem-specific
background theory.

\section{Framework}
\label{Framework}

In the following, we assume familiarity with the logic programming terminology
introduced e.g. in \cite{Lloyd1987}.

\subsection{Meta-Interpretive Learning}
\label{Meta-Interpretive Learning}

\begin{table}[t]
	\small
	\centering
	\begin{tabular}{ll}
		\textbf{Name}    & \textbf{Metarule} \\
		\toprule
		\emph{Identity}  & $\exists P,Q    \; \forall x,y$:     $P(x,y) \leftarrow Q(x,y)$\\ 
		\emph{Inverse}   & $\exists P,Q    \; \forall x,y$:     $P(x,y) \leftarrow Q(y,x)$\\ 
		\emph{Precon}    & $\exists P,Q    \; \forall x,y$:     $P(x,y) \leftarrow Q(x), R(x,y)$\\ 
		\emph{Chain}     & $\exists P,Q,R  \; \forall x,y,z$:   $P(x,y) \leftarrow Q(x,z), R(z,y)$\\ 
		\bottomrule
	\end{tabular}
\caption{Metarules commonly found in the MIL literature.}
\label{tab:mil_metarules}
\end{table}

Meta-Interpretive Learning\cite{Muggleton2014,Muggleton2015} (MIL) is a setting
for ILP where first-order logic theories are learned by SLD-Refutation of ground
atoms $E^+, E^-$ given as positive and negative training examples, respectively.
$E^+, E^-$ are resolved with a higher-order background theory that includes both
first- and secod-order definite clauses, $B$ and $\mathcal{M}$, respectively,
the latter known as ``metarules"\cite{Muggleton2015}. A first SLD-Refutation
proof of $E^+$ produces an initial hypothesis that is then specialised by a
second SLD-Refutation step, this time proving $E^-$ to identify, and discard,
over-general hypotheses. The ``metarules", are second-order definite clauses
with second-order variables existentially quantified over the set of predicate
symbols, which can optionally include automatically generated invented predicate
symbols, and with first-order variables existentially or universally quantified
over the set of first-order terms, in $B,E^+$. During SLD-Refutation of $E^+$
metarules' second-order variables are unified to predicate symbols, constructing
the clauses of a first-order hypothesis, $H$. Table \ref{tab:mil_metarules}
lists metarules commonly found in the MIL literature. 

\subsection{A Self-Supervised Learning setting for ILP}
\label{A Self-Supervised Learning setting for ILP}

In the Normal Setting for ILP\cite{Nienhuys-Cheng1997}, we are given a
first-order background theory $B$, positive and negative examples $E^+, E^-$,
and must find a \emph{correct hypothesis}, a logic program $H$, such that $B
\cup H \models E^+$ and $\forall e^- \in E^-: B \cup H \not \models e^-$. We now
formally define our new, Self-Supervised Learning setting for ILP,
\emph{SS-ILP}, where there are no negative examples, only labelled positive, or
unlabelled examples, and $B$ is replaced by a higher-order background theory
that is maximally general. We model our definition after the definition of the
Normal Setting for ILP formalised in \cite{Nienhuys-Cheng1997}.

\begin{definition} \textbf{Inductive Logic Programming: Self\-/Supervised Setting}
	\begin{itemize}
		\item \textbf{Not Given:}

	$\mathbf{\Theta} = P/n$, a predicate $P$ of arity $n$ that we call the
	\emph{target predicate}. Let $B_H$ be the Herbrand Base of $\Theta$.
	Recall that the Herbrand Base of a predicate $P/n$ is the set of all
	ground atoms $P/n$ with variables substituted for ground terms in the
	Domain of Discourse, $\mathcal{U}$, (or a new constant $\alpha$ if there
	are no ground terms in $\mathcal{U}$) \cite{Nienhuys-Cheng1997}. 

	$\mathbf{I}$, a Herbrand interpretation over $B_H$ that we call the
	\emph{intended interpretation}. Let $I^+, I^- \subseteq B_H$ be the sets
	of atoms in $B_H$ that are true and false under $I$, respectively, such
	that $I^+ \cup I^- = B_H$ and $I^+ \cap I^- = \emptyset$. 

	\item \textbf{Given:}

	$\mathbf{E = \{E^+,E^?\} \subseteq B_H}$, a finite set of ground atoms
	$P/n$. Let $E^+ \subseteq I^+$, and $E^? \subseteq I^+ \cup I ^-$.
	Either $E^+$ or $E^?$ may be empty, but not both. We say that examples
	in $E^+$ are \emph{labelled} and examples in $E^?$ are
	\emph{unlabelled}, denoting an assumption that $E^+$ are all true under
	$I$ and uncertainty of which atoms in $E^?$ are true under $I$.

	$\mathbf{\mathcal{T} = B \cup \mathcal{M}}$, a higher-order background
	theory, where $B,\mathcal{M}$ are finite sets of first- and second-order
	definite clauses, respectively, and such that $\mathcal{M} \cup B \cup E
	\models I^+ \cup I^-$, i.e. $\mathcal{T}$ is maximally general with
	respect to $\Theta$.

	\item \textbf{Find:}

	$\mathbf{H}$, a logic program that we call a \emph{hypothesis}, a finite
	set of first-order definite program clauses such that:
	
	\begin{enumerate}
		\item $B \cup \mathcal{M} \models H$
		\item $B \cup H \models I^+$ 
		\item $\forall a \in I^-: B \cup H \not \models a$
	\end{enumerate}

	$\mathbf{L = L^+ \cup L^-}$, a set of two finite sets of atoms that we
	call a \emph{labelling}, such that:

	\begin{enumerate}
		\item $L^+ \subseteq I^+$
		\item $L^- \subseteq I^-$
	\end{enumerate}

	If $H, L^+, L^-$ satisfy the above criteria we call $H$ a \emph{correct
	hypothesis} and $L$ a \emph{correct labelling} under the Self-Supervised
	setting for ILP.
	\end{itemize}

	\label{def:ssl_ilp_setting}
\end{definition}


\subsection{Poker: an algorithm for Self-Supervised ILP}
\label{Poker: a Self-Supervised algorithm for ILP}

Poker assumes that the higher-order background theory, $\mathcal{T}$, is a
\emph{Second-Order Definite Normal Form}, abbreviated \emph{SONF} for ease of
pronounciation. Informally, a SONF is a set of \emph{constrained metarules} that
generalise the set of first-order logic program definitions of \emph{all
possible interpretations $I$} over the Herbrand Base $B_H$ of $\Theta$. A formal
definition of constrained metarules and SONFs, follows.

First, we define constrained metarules extending the definition of metarules in
\cite{Muggleton2015} with a set of constraints on the substitution of
second-order variables.

\begin{definition} [Constrained Metarules] Let $B,\mathcal{M},E^+$ be as in
	Definition \ref{def:ssl_ilp_setting}, and let $\mathcal{P}$ be the set
	of all predicate symbols in $B, E^+$, and 0 or more automatically
	generated invented predicate symbols. Let $S = \{S_E, S_B, S_I,
	S_{\epsilon}\}$ be a set of disjoint subsets of $\mathcal{P}$, the sets
	of symbols in $E^+$, $B$, invented predicates, and the set $\{\epsilon\}
	\subseteq S_B$, respectively, and let $P, Q$ be two distinct
	second-order variables, and $P_1,...,P_n$ a list of $n$ such, in a
	metarule $M \in \mathcal{M}$. A \emph{metarule constraint} on $M$ is one
	of the following atomic formulae, with the associated interpretation:
	\begin{itemize}
		\item $P = Q$: $P,Q$ \emph{must} be instanted to the same symbol
			in $\mathcal{P}$.
		\item $P \neq Q$: $P,Q$ \emph{must not} be instantiated to the
			same symbol in $\mathcal{P}$.
		\item $target(P_1,...,P_n)$: each $P_i$ must be instantiated to
			a symbol in $S_E$.
		\item $invented(P_1,...,P_n)$: each $P_i$ must be instiantiated
			to a symbol in $S_I$.
		\item $background(P_1,...,P_n)$: each $P_i$ must be
			instantiated to a symbol in $S_B$.
		\item $empty(P_1,...,P_n)$ : $P$ must be instantiated to
			$\epsilon$.
		\item $invented(P,Q) \rightarrow P \neq Q$: If $P,Q$ are
			instantiated to symbols in $S_I$, $P$ must not be the
			same symbol as $Q$.
		\item $invented(P,Q) \rightarrow P \geq Q$: If $P,Q$ are
			instantiated to symbols in $S_I$, $P$ must be above than
			or equal to $Q$ in the standard alphabetic ordering.
		\item $invented(P,Q) \rightarrow P \leq Q$: If $P,Q$ are
			instantiated to symbols in $S_I$, $P$ must be below than
			or equal to $Q$ in the standard alphabetic ordering. 
	\end{itemize}

	If $C_M^1, C_M^2$ are two constraints on metarule $M$, then $C_M^1
	\wedge C_M^2$ is a constraint on $M$, interpreted as "$C_M^1$ and
	$C^2_M$ must both be true".

	If $C_M^1, C_M^2$ are two constraints on metarule $M$, then $C_M^1 \vee
	C_M^2$ is a constraint on $M$, interpreted as "one or both of $C_M^1$,
	$C^2_M$ must be true".

	If $C_M$ is a constraint on metarule $M$, then $\neg C_M$ is a
	constraint on $M$, interpreted as the opposite of $C_M$. E.g. $\neg
	target(P)$ is interpreted as "$P$ must not be instantiated to a symbol
	in $S_E$".

	A \emph{constrained metarule} is a formula $M : C_M$, where $M \in
	\mathcal{M}$ and $C_M$ is a constraint on $M$.

	Constraints on a metarule $M$ are satisfied by a first order definite
	clause $C$ if, and only if, $C$ is an instance of $M$ with second-order
	variables instantiated according to the constraints on $M$.
	\label{def:constrained_metarules}
\end{definition} 

Metarule constraints are primarily meant to control recursion, particularly to
eliminate unnecessary recursion, including left recursion, and to reduce
redundancy in the construction of Poker's initial hypotheses. Metarule
constraints can thus improve Poker's efficiency, and in that sense serve much
the same function as heuristics exploiting the common structure of problems in a
class, as e.g. in Planning \cite{GeffnerAndBonet} and SAT-Solving
\cite{Biere2021}.

We now formalise the definition of SONFs.

\begin{definition} [Second-Order Definite Normal Form] Let $\Theta$, $B_H$, $I$,
	$I^+$, $I^-$, and $\mathcal{M}$ be as in Definition
	\ref{def:ssl_ilp_setting}, with the additional stipulation that
	$\mathcal{M}$ is a set of \emph{constrained} metarules as in Definition
	\ref{def:constrained_metarules}. Let $\mathcal{I}$ be the set of all
	Herbrand interpretations $I$ over $B_H$ and, for each $I \in
	\mathcal{I}$ let $\Pi_I$ be the set of all first-order definite programs
	$\Sigma_1, ..., \Sigma_n$, such that $\forall \Sigma_i \in \Pi_I:
	\Sigma_i \models I^+$ and $\forall \Sigma_i \in \Pi_I, \forall e^- \in
	I^-: \Sigma_i \not \models e^-$. Let $\Pi^*$ be the set of all $\Pi_I$
	for all $I$ over $B_H$.

	$\mathcal{M}$ is a Second-order Definite Normal Form, abbreviated
	to \emph{SONF}, for $\Theta$, if, and only if, $\mathcal{M} \models
	\Pi'$ for some subset $\Pi'$ of $\Pi^*$. $\mathcal{M}$ is a \emph{strong
	SONF} for $\Theta$ if $\Pi' = \Pi^*$. $\mathcal{M}$ is a \emph{weak
	SONF} for $\Theta$ if $\Pi' \subset \Pi^*$ (i.e.  $\Pi'$ is a subset of,
	but not equal to, $\Pi^*$).

	\label{def:normal_forms}
\end{definition}


The purpose of SONFs is to replace the use of problem-specific sets of
metarules, previously common in MIL practice, with a maximally general
second-order background theory sufficient to learn any logic program definition
of a target predicate $\Theta$, for any interpretation $I$ of $\Theta$.

\begin{table}[t]
	\small
	\centering
	\begin{tabularx}{\columnwidth}{l}
		\multicolumn{1}{c}{ \textbf{Chomsky-Greibach Second-Order Definite Normal Form} }\\
		\toprule
		\textbf{(Identity)} $P(x,y) \leftarrow Q(x,y):$ \\
		$target(P) $ $\wedge \; background(Q) \vee empty(Q) $ \\
		\midrule
		\textbf{(Chain)} $P(x,y) \leftarrow Q(x,z), R(z,y):$ \\
		$P \neq Q $ $\wedge \; ( target(P) \vee invented(P) ) $ $\wedge \; not(target(Q)) $ \\
		$\wedge \; not(empty(Q,R)) $ $\wedge \; invented(P,Q) \rightarrow P \neq Q $\\
		\midrule
		\textbf{(Tri-Chain)} $P(x,y) \leftarrow Q(x,z), R(z,u),	S(u,y):$ \\
		$ P \neq Q$ $\wedge \; Q \neq R$ $\wedge \; R \neq S$\\
		$\wedge \; ( target(P) \vee invented(P)$ ) $\wedge \; not(target(Q) $ \\
		$\wedge \; not(empty(Q,R,S))$ $\wedge \; invented(P,Q) \rightarrow P \neq Q $\\
	\bottomrule
	\end{tabularx}
	\caption{Chomsky-Greibach SONF for CFL DCGs.}
	\label{tab:full_cgnf}
\end{table}

\begin{table*}[t]
	\small
	\centering
	\begin{tabular}{llll}
		\textbf{Chomsky Normal Form} & \textbf{DCG} & \textbf{Definite Clauses} & \textbf{C-GNF metarules} \\
		\toprule
		$s \rightarrow \epsilon$ & $s \rightarrow empty.$    & $s(x,y) \leftarrow empty(x,y).$           & \textbf{Identity} $P(x,y) \leftarrow Q(x,y)$ \\
		$N_0 \rightarrow N_1N_2$ & $n_0 \rightarrow n_1,n_2. $ & $n_0(x,y) \leftarrow n_1(x,z), n_2(z,y).$ & \textbf{Chain} $P(x,y) \leftarrow Q(x,z), R(z,y)$ \\
		$N_i \rightarrow t$      & $n_i \rightarrow t.$      & $n_i(x,y) \leftarrow t(x,y).$             & \textbf{Identity} $P(x,y) \leftarrow Q(x,y)$ \\
					 & $empty \rightarrow [\;].$   & $empty(x,x)$ 				 & None (included in $B$) \\
					 & $t \rightarrow [t].$	     & $t([t|x],x)$ 				 & None (included in $B$) \\
		\bottomrule
	\end{tabular}
	\caption{Chomsky Normal Form mapping to DCGs and C-GNF constrained
	metarules. $N_i, n_i$: non-terminals; $t$ : terminals.}
	\label{tab:cnf}
\end{table*}

\subsubsection{Two Normal Forms}

We illustrate the concept of Second Order Definite Normal Forms with two SONFs
used in the Experiments Section: Chomsky-Greibach Normal Form (C-GNF) used to
learn CFGs, and Lindenmayer Normal Form (LNF) used to learn L-System grammars.
C-GNF is listed in Table \ref{tab:full_cgnf}. We list LNF and prove the
completeness of C-GNF for CFLs, and LNF for L-systems, in the Appendix.

Derivation of a SONF is a process of abstraction that we do not currently know
how to automate. Both C-GNF and LNF are therefore derived manually by the
author. C-GNF is derived by an encoding of the rules of Chomsky and Greibach
Normal Forms in our Second-Order Normal Form notation. LNF is derived from the
common structure observed in manual definitions of L-System grammars as DCGs,
coded by the author. Metarule constraints are obtained from experimentation to
improve learning efficiency.

A guide for the construction of Second-Order Normal Forms is beyond the scope of
the paper. However, we include Table \ref{tab:cnf} as an illustration of the
derivation of C-GNF metarules (without constraints) from Chomsky Normal Form. 

\subsubsection{Poker's algorithm}

We typeset Poker's algorithm for SS-ILP as Algorithm \ref{alg:poker}. We state
our main theoretical result as Theorem \ref{thm:poker}. We include a proof of
the Theorem in the Appendix.


\begin{algorithm}[t]
	\small
	\caption{Poker: SS-ILP by detection of contradictions}
	\label{alg:poker}
	\textbf{Input:} $E^+,E^?,B,\mathcal{M}$ as in Definition \ref{def:ssl_ilp_setting}; $\mathcal{M}$ is a SONF.\\
	$V$, finite set of invented predicate symbols.\\
	Integer $k \geq 0$, max. number of automatically generated examples. \\
	Integer $l \geq 0$, max. number of clauses in initial hypotheses $H$.\\
	\textbf{Initialise:} Hypothesis $T \leftarrow \emptyset$. Labelling $L = \{E^+, E^- \leftarrow \emptyset\}$.
	\begin{algorithmic}[1]
		\Procedure{Generalise}{$B,\mathcal{M},E^+$}
			\State $T \leftarrow \{H_{|H| \leq l} : \mathcal{M} \cup B \cup V \models H \wedge B \cup H \models e^+ \in E^+\}$ 
		\EndProcedure
		\Procedure{Generate}{$B,T,E^?$}
			\State $T' \leftarrow \bigcup T$ \Comment This step is optional. Alternatively $T' \leftarrow T$  \label{alg:step_gestalt}
			\State $E^- \leftarrow E^? \cup \{e_{i=1}^k : \; B \cup T' \models e_i\}$ 
		\EndProcedure
		\Procedure{Label}{$B,T,E^+,E^-$}
			\ForAll{$e \in E^-$} 
				\State $T' \leftarrow T \setminus \{H : \; H \in T \wedge B \cup H \models e\}$ 
				\If{$B \cup T' \models E^+$}
					\State $T \leftarrow T'$
				\Else
					\State $E^- \leftarrow E^- \setminus \{e\}$ 
					\State $E^+ \leftarrow E^+ \cup \{e\}$
				\EndIf
			\EndFor
		\EndProcedure
		\State $T \leftarrow \bigcup T$ \Comment This step is optional.
		\State $T \leftarrow T \setminus \{C \in T: T \models C\} $ \Comment This step is optional.
		\State \textbf{Return} $T, L = \{E^+, E^-\}$
	\end{algorithmic}
\end{algorithm}

\begin{theorem} [Hypothesis Correctness] The probability that Algorithm
	\ref{alg:poker} returns a correct hypothesis increases monotonically
	with the number of unlabelled examples.
	\label{thm:poker}
\end{theorem}

\section{Implementation}
\label{Implementation}

We implement Algorithm \ref{alg:poker} in a new system, also called Poker,
written in Prolog. Poker extends the Top Program Construction algorithm
\cite{Patsantzis2021a} (TPC). Procedure \textsc{Generalise} in Algorithm
\ref{alg:poker}, taken from TPC, is implemented in Poker by a call to Vanilla,
an inductive second-order Prolog meta-interpreter for MIL, introduced in
\cite{Trewern24}. Vanilla is developed with SWI-Prolog
\cite{wielemaker:2011:tplp} and uses SWI-Prolog's tabled execution,
\cite{TamakiAndSato1986}, a.k.a. SLG-Resolution, to control recursion and
improve performance. 

\section{Experiments}
\label{Experiments}

Poker learns from three kinds of examples: user-given a) labelled, and b)
unlabelled, examples, and, c) automatically generated examples. In preliminary
experiments we find that the effect of (b), unlabelled examples, on Pokers'
learning performance depends on the generality of the unlabelled examples. This
merits more thorough investigation. We limit the current investigation on the
effect of automaticaly generated examples on Poker's performance, formalised in
the following research question:

\begin{question} What is the effect of automatically generated examples on
	Poker's learning performance?
\end{question}

We address this question with two sets of experiments, learning grammars for two
sets of languages: Context-Free Languages (CFLs); and L-Systems. L-System
grammars are meant to be run as \emph{generators}, producing strings interpreted
as movement and drawing commands sent to a (now usually simulated) robot.
Accordingly, we evaluate learned L-System grammars as \emph{generators}.

\subsubsection{Baselines} There are, to our knowledge, no ILP systems that
generate new examples during learning, other than Poker. We consider comparing
Poker to ILP systems Aleph \cite{Srinivasan2001} and Popper \cite{Cropper2021}.
In preliminary experiments both perform poorly without negative examples.
In private communication Popper's authors confirm Popper should not be expected
to learn CFGs without negative examples. We alight on the state-of-the-art MIL
system Vanilla-Louise \cite{Trewern24} (hereby denoted as Louise, for brevity),
as a simple baseline. Louise is also based on Vanilla and accepts constrained
metarules, simplifying comparison.



\subsection{Experimental protocol}
\label{Experimental protocol}

\begin{table}[t]
	\setlength{\tabcolsep}{1mm}
	\small
	\centering
	\begin{tabularx}{\columnwidth}{llll}
		\multicolumn{4}{c}{ \textbf{Background theories} } \\
		\toprule
		\multicolumn{2}{c}{ \textbf{Low Uncertainty (CFLs)} } & \multicolumn{2}{c}{ \textbf{Mod. Uncertainty (L-Systems)} }  \\
		\midrule
		\textbf{DCG Rule} & \textbf{Definite Clause}       &	\textbf{DCG Rule}            & \textbf{Definite Clause} \\ 
		$one \rightarrow [1]$    & $one([1|X],X)$      & $plus \rightarrow [+]$  & $plus([+|X],X)$           \\
		$zero \rightarrow [0]$   & $zero([0|X],X)$     & $min \rightarrow [-]$ & $min([-|X],X)$          \\
		$empty \rightarrow [\;]$ & $empty(X,X)$        & $f \rightarrow [f]$     & $f([f|X],X)$              \\
					 & 		       & $g \rightarrow [g]$     & $g([g|X],X)$              \\
					 & 		       & $x \rightarrow [x]$     & $x([x|X],X)$              \\
					 & 		       & $y \rightarrow [y]$     & $y([y|X],X)$              \\
		\bottomrule
	\end{tabularx}                                         
		\caption{Background theories used in experiments.}
		\label{tab:bk}
\end{table}

In each set of experiments, we generate positive training examples and positive
and negative testing examples from a manual definition of the learning target
(the grammar to be learned) as a DCG.

We train Poker and Louise on increasing numbers, $l > 0$, of positive labelled
examples. We vary the number of Poker's automatically generated examples, $k
\geq 0$, and record, a) for CFLs, the True Positive Rate (TPR) and True Negative
Rate (TNR) of i) hypotheses learned by both systems and ii) labellings returned
by Poker; and b) for L-Systems i) the accuracy of learned hypotheses as
generators (``Generative Accuracy") and ii) the cardinality of learned
hypotheses, i.e. their number of clauses (``Hypothesis Size"). Testing examples
are used to evaluate CFL hypotheses as acceptors: for each hypothesis, $H$, we
test how many positive testing examples are accepted, and how many negative
ones rejected, by $H$. We do not use testing examples to evaluate labellings or
L-System hypotheses as generators. Instead, we count the number of examples
labelled positive or negative, in a labelling, or generated by a hypothesis
executed as a generator, that are accepted or rejected, respectively for
positive and negative examples, by the manually defined learning target.

In CFL experiments, we measure TPR and TNR to avoid confusing measurements when
$k = 0$. 

\begin{figure*}[t]
	\centering
	\includegraphics[width=0.85\textwidth]{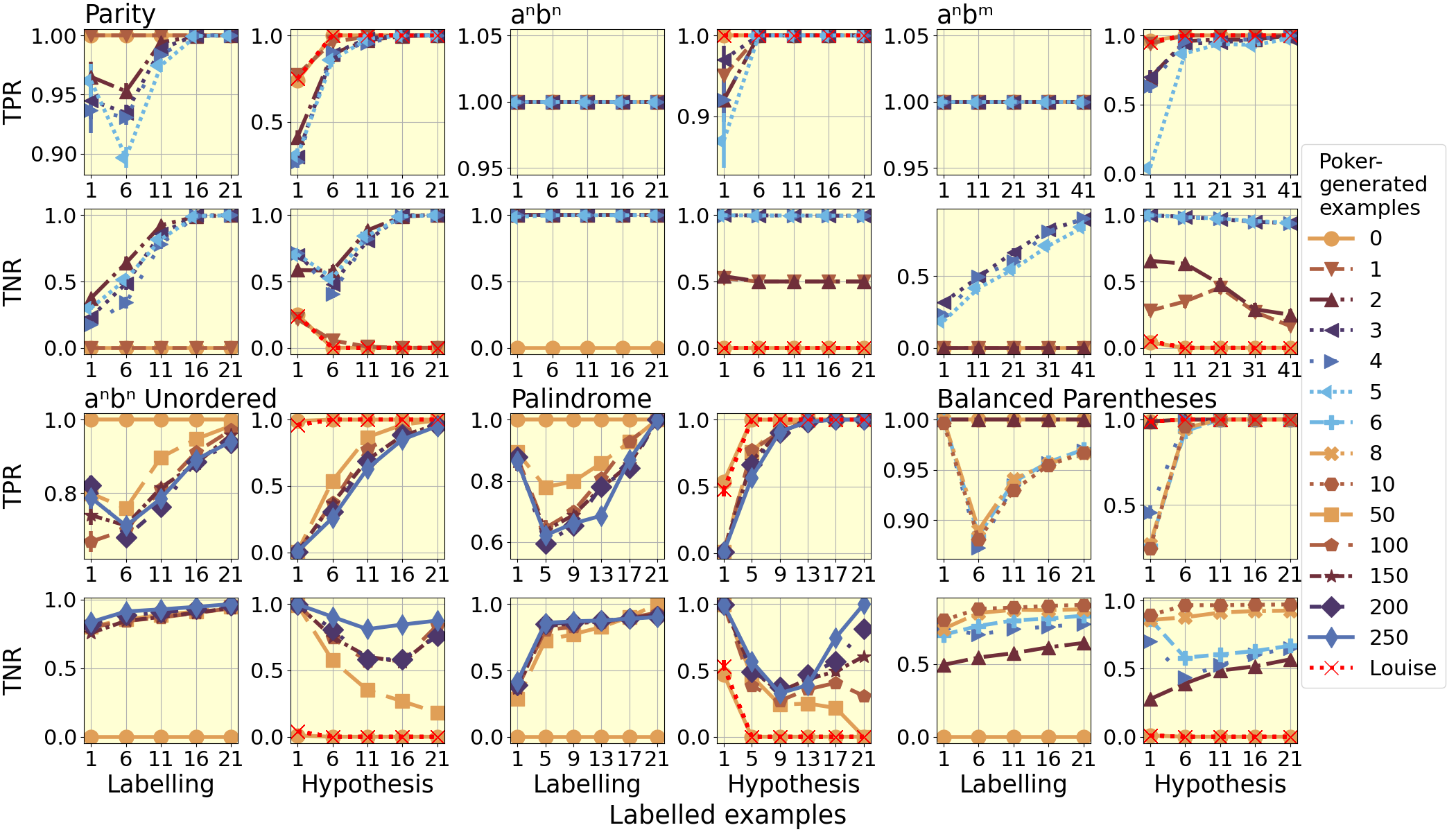}
	\caption{Learning CFGs. TPR: True Positive Rate. TNR: True Negative
	Rate.
	}
	\label{fig:low_uncertainty}
\end{figure*}

Experiment results are listed in Figures \ref{fig:l_system_generator} and
\ref{fig:low_uncertainty} for L-Systems and  CFLs, respectively. In each Figure,
the x-axis plots the number of labelled positive examples, the y-axis plots mean
TPR, TNR, Generative Accuracy or Hypothesis Size, while each line plots the
change in that metric with each increment of $k$, the number of automatically
generated examples. For Louise, $k = 0$ always. Error bars are standard error
over 100 randomly sampled sets of training and testing examples in each
experiment.

\subsection{Experiment 1: L-Systems}
\label{Experiment 1: L-Systems}

\begin{figure}[!htbp]
	\centering
	\includegraphics[width=0.45\textwidth]{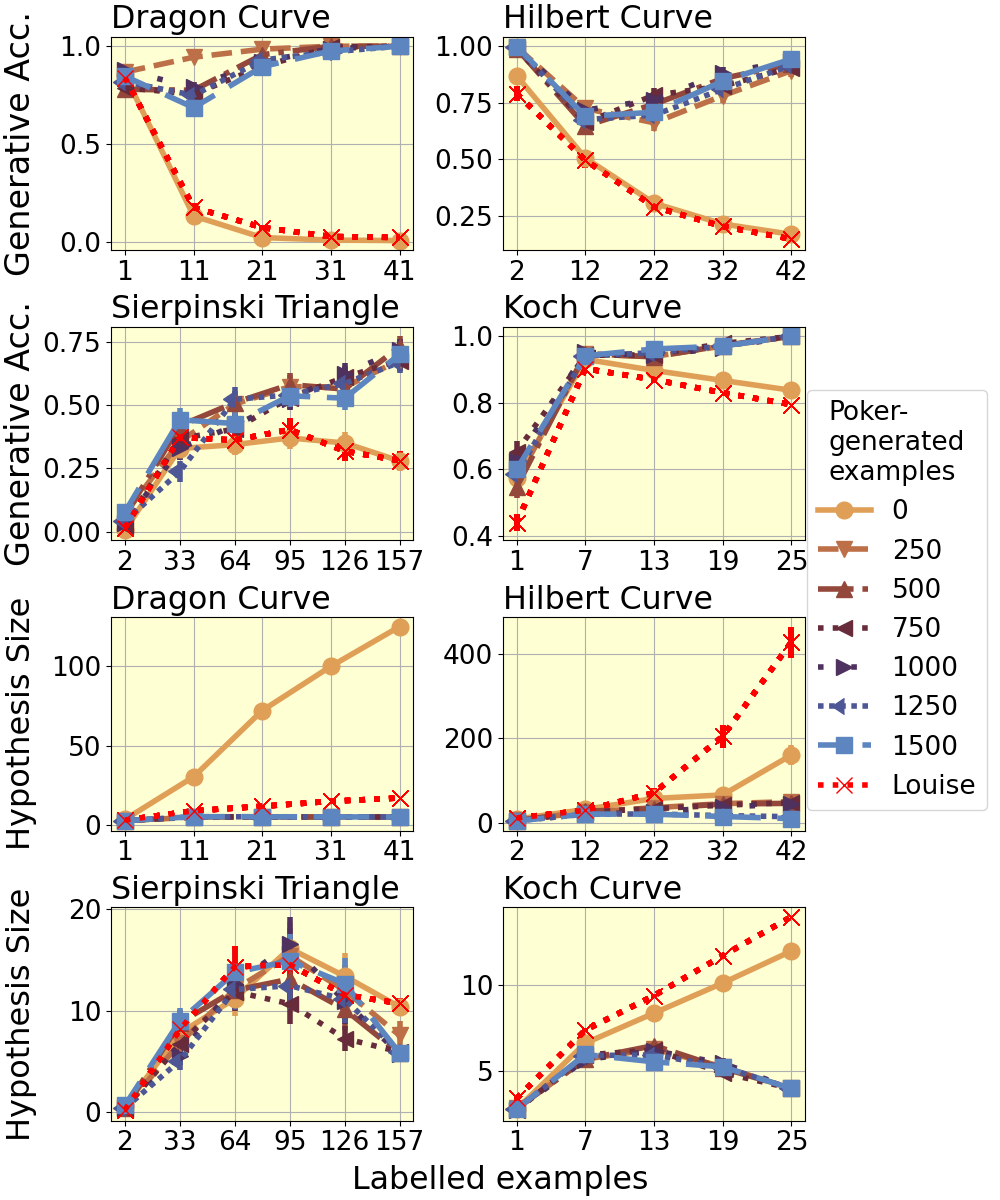}
	\caption{Learning L-Systems: generative accuracy.
	}
	\label{fig:l_system_generator}
\end{figure}

We train Poker and Louise on example strings of the L-Systems for the Dragon
Curve, Hilbert Curve, Koch Curve, and Sierpinski Triangle fractals, taken from
\cite{ABoP1996}. The first-order background theory consists of a set of symbols,
common for all L-Systems, but not necessarily with the same meaning for each,
e.g. the symbol $f$ is a \emph{variable} (nonterminal) in Dragon and Koch Curve,
but a \emph{constant} (terminal) in Hilbert Curve. This serves as a test of a
system's robustness to noise. The second-order background theory is LNF, for all
L-systems.

Results are shown in Figure \ref{fig:l_system_generator}. We observe that
Poker's Generative Accuracy increases, and Hypothesis Size decreases, with the
number of automatically generated examples. Louise's Generative Accuracy
decreases and Hypothesis Size increases with the number of labelled examples,
indicating increasing over-generalisation. This indicates that increasing
numbers of automatically generated examples improve Poker's performance and
avoid over-generalisation in language generation tasks.

\subsection{Experiment 2: Binary CFLs}
\label{Experiment 2: Binary CFLs}

We train Poker and Louise on examples of six CFLs: a) The language of
bit-strings with an even number of 1's (even parity), b) $\{a^nb^n: n > 0 \}$,
c) $\{w \in \{a,b\}^*: n_a(w) = n_b(w), n \geq 0\}$ (n a's and n b's in any
order), d) $\{a^nb^m|n \geq m \geq 0\}$, e) the language of palindromic
bit-strings and f) the language of balanced parentheses. Symbols $a,b,)$, and
$($ are suitably replaced by 1 and 0, so that all experiments can share the same
first- and second-order background theory, consisting of the alphabet
$\{1,0,\epsilon\}$, in DCG form, as listed in Table \ref{tab:bk}, and
C-GNF\footnote{Note that the Tri-Chain metarule is only used with Palindrome.},
respectively.

Results are listed in Figure \ref{fig:low_uncertainty}. When $k = 0$, Poker's
TPR is maximal while its TNR is minimal, because there are no negative examples.
When $k > 0$ both TPR and TNR increase with $k$, until they are both maximised
with the highest increments of $k$. From this we conclude that when $k$ is low,
Poker's labelling and hypothesis over-generalise, and when $k$ is sufficiently
large, Poker learns a correct labelling and hypothesis. Thus, Poker's
performance improves with the number of automatically generated examples.
Louise over-generalises consistently over all experiments.

\section{Conclusions and Future Work}
\label{Conclusions and Future Work}

ILP systems can generalise well from few examples given a problem-specific
background theory and negative examples, both manually selected with domain
knowledge. We have endeavoured to address this onerous requirment with a new
formal setting for Self-Supervised ILP, SS-ILP, and presented Poker, a new MIL
system that learns in the new setting. Poker's algorithm learns from labelled
and unlabelled examples, and automatically generates new positive and negative
examples during learning. Instead of a problem-specific background theory Poker
only needs a maximally general, Second Order Normal Form (SONF). We have
presented two SONFs, for Context-Free and L-System Definite Clause Grammars. We
have given a theoretical proof and empirical evidence that Poker's accuracy
improves with increasing numbers of unlabelled examples, and showed how a
baseline MIL system lacking Poker's ability to automatically generate negative
examples over-generalises.

Our theoretical results can be extended with further proofs of the correctness
of Poker's algorithm, including with respect to varying amounts of labelled and
unlabelled examples; and of its computational efficiency. Our empirical results
are restricted to grammar learning. Future work should investigate Poker's
application to more diverse domains.

\section{Acknowledgments} We thank Em. Professor Stephen Muggleton and Drs Lun
Ai and C\'eline Hocquette for their kind feedback on this work.

\bibliography{mybib}

\appendix

\newpage

\section{Proof of Theorem 1}

In this section we prove our main theoretical result in Theorem \ref{thm:poker}.

Lemma \ref{lem:one} states that all correct hypotheses are included in the set
of initial hypotheses.

\begin{lemma} [Generality of Initial Hypotheses] Let $\Theta$, $I^+$, $I^-$ be
	as in Definition \ref{def:ssl_ilp_setting}. Let $B$, $\mathcal{M}$,$
	E^+$, $V$, $l$, be as in the inputs of Algorithm \ref{alg:poker} where
	$\mathcal{M}$ is a Second-Order Definite Normal Form for $\Theta$ and
	$E^+ \subseteq I^+$. Let $H^*$ be a correct hypothesis according to
	Definition \ref{def:ssl_ilp_setting} such that $\mathcal{M} \cup B \cup
	V \models H^*$, $B \cup H^* \models E^+$, and $|H^*| \leq l$. And let
	$T$ be the set of initial hypotheses constructed by Procedure
	\textsc{Generalise} in Line 10 of Algorithm \ref{alg:poker}. Then,
	$\forall H^*: H^* \in T$.
	\label{lem:one}
\end{lemma}

\begin{proof} Follows from the refutation completeness of Second-Order
	SLD-Resolution \cite{Trewern24} and the definition of Second-Order
	Definite Normal Forms in Definition \ref{def:normal_forms}. Because of
	the refutation completeness of Second-Order SLD-Resolution, Procedure
	\textsc{Generalise} derives each hypothesis $H$ such that $\mathcal{M}
	\cup B \cup V \models H$ and $B \cup H \models E^+$ where $|H| \leq l$,
	in Line 2 of Algorithm \ref{alg:poker}. This is true for each $H = H^*$
	because if $\mathcal{M}$ is a Second Order Definite Normal Form for
	$\Theta$, then $\mathcal{M} \models H^*$, according to Definition
	\ref{def:normal_forms}, therefore $\mathcal{M} \cup B \cup V \models
	H^*$ also, and $B \cup H^* \models E^+$ by the assumption in Lemma
	\ref{lem:one}.
\end{proof}

Lemma \ref{lem:two} states that all hypotheses accepting a negative example are
removed from the set of initial hypotheses.

\begin{lemma} [Correct Specialisation] Let $B$, $T$, $E^+$, $E^-$ be as in the
	inputs of Procedure \textsc{Label} in Algorithm \ref{alg:poker} and let
	$T' = T \setminus \{H : \; H \in T \wedge B \cup H \models e\}$, the set
	constructed by Procedure \textsc{Label} in Line 10 of Algorithm
	\ref{alg:poker}. Then, $\forall e \in E^-: T' \not \models e$.
	\label{lem:two}
\end{lemma}

\begin{proof} Follows from the soundness of (first-order) SLD-Resolution
	\cite{Nienhuys-Cheng1997}. If $T'$ is the set $T$, set-minus all $H \in
	T$ such that $B \cup H \models e$, then because of the soundness of
	SLD-Resolution each such $H$ is correctly removed from $T$ on Line 10 of
	Algorithm \ref{alg:poker} and there remains no H in $T'$ such that $B
	\cup H \models e$.
\end{proof}

Lemma \ref{lem:three} states that only hypotheses accepting negative examples
are removed from the set of initial hypotheses.


\begin{lemma} [Correct Labelling] Let $e \in E^-$ be a negative example, i.e. $e
	\in I^-$, let $T, T'$ be as in Lemma \ref{lem:two}, and let $T'' = \{H :
	\; H \in T \wedge B \cup H \models e\}$ be the set of hypotheses removed
	from the set of initial hypotheses $T$, by Procedure \textsc{Label} in
	Line 10 of Algorithm \ref{alg:poker}. Then, for each hypothesis $H'' \in
	T''$, $H''$ is not a correct hypothesis, according to Definition
	\ref{def:ssl_ilp_setting}.
	\label{lem:three}
\end{lemma}

\begin{proof} Follows from Lemma \ref{lem:two} and the definition of correct
	hypotheses in Definition \ref{def:ssl_ilp_setting}. If $e \in I^-$ then
	according to Lemma \ref{lem:two}, for each hypothesis $H''$ in $T''$, it
	is true that $B \cup H'' \models e$, therefore it is not true that
	$\forall a \in I^-: B \cup H'' \not \models a$, and $H''$ is not a
	correct hypothesis according to Definition \ref{def:ssl_ilp_setting}.
\end{proof}

Lemma \ref{lem:four} states the least number of negative examples needed to
remove all hypotheses that accept negative examples from the set of initial
hypotheses.

\begin{lemma} [Least Negative Examples] Let $n$ be the number of all hypotheses,
	and let $m$ be the number of all correct hypotheses, in the set $T$ of
	initial hypotheses constructed by Procedure \textsc{Generalise} in Line
	2 of Algorithm \ref{alg:poker}. Procedure \textsc{Label} in Line 10 of
	Algorithm \ref{alg:poker} constructs the set $T'' \subseteq T$ of all
	hypotheses $H'' \in T$ such that $B \cup H'' \models e$, for each
	unlabelled example $e \in E^-$. The least cardinality of the set $E^-$
	sufficient to reduce $T$ to a set of only correct hypotheses is $n - m$.
	\label{lem:four}
\end{lemma}

\begin{proof} Follows from Lemma \ref{lem:three}. For each hypothesis $H''$ in
	the set $T$ of initial hypotheses, where $H''$ is not a correct
	hypothesis, Procedure \textsc{Label} removes $H''$ from $T$ given any
	negative example $e \in E^-$ such that $B \cup H'' \models e$, therefore
	a single negative example is sufficient to remove each such $H''$. The
	number of such $H''$ is $n - m$, therefore $n - m$ such examples are
	sufficient to remove each such $H''$.
\end{proof}

We now restate, and prove, Theorem \ref{thm:poker}.

\begin{theorem} [Hypothesis Correctness] The probability that Algorithm
	\ref{alg:poker} returns a correct hypothesis increases monotonically
	with the number of unlabelled examples.
\end{theorem}

\begin{proof} Follows from Lemmas \ref{lem:one} - \ref{lem:four}. According to
	Lemma \ref{lem:one}, because of the refutation completeness of
	Second-Order SLD-Resolution the set of initial hypotheses $T$ includes
	one or more correct hypotheses $H^*$. It further follows from the
	refutation completeness of Second-Order SLD Resolution that $T$ also
	includes zero or more hypotheses $H''$ which are not correct hypotheses.
	We shal refer to hypotheses other than correct hypotheses as
	\emph{incorrect hypotheses} for the purpose of this proof. According to
	Lemmas \ref{lem:two} - \ref{lem:three}, Procedure \textsc{Label} removes
	only each incorrect hypothesis $H''$ in $T$ such that $B \cup H''
	\models e$ for each unlabelled example $e \in E^-$ where $e$ is a
	negative example, i.e. $e \in I^-$ by Definition
	\ref{def:ssl_ilp_setting}. According to Lemma \ref{lem:four} the number
	of incorrect hypotheses $H''$ that must be removed from $T$ to leave
	behind only correct hypotheses is $n - m$, where $n = |T|$ and $m$ is
	the number of correct hypotheses $H^* \in T$.  As the cardinality of
	$E^-$ increases, the chance that it includes a negative example
	increases. With each negative example in $E^-$ the number of negative
	examples in $E^-$ increases towards $n - m$, sufficient to remove each
	incorrect hypothesis $H''$ from $T$. No new hypotheses are added to $T$
	after $T$ is constructed by Procedure \textsc{Generalise} on line 2 of
	Algorithm \ref{alg:poker}, therefore the number $n - m$ of incorrect
	hypothesers $H''$ in $T$ never increases, instead it decreases
	monotonically, approaching 0. As the number of incorrect hypotheses
	decreases monotonically the probability that $T$ includes only correct
	hypotheses, therefore that Algorithm \ref{alg:poker} returns a correct
	hypothesis increases monotonically.
\end{proof}

\newpage

\section{Chomsky-Greibach Second-Order Definite Normal Form}

\begin{table}[!htbp]
	\small
	\centering
	\begin{tabularx}{\columnwidth}{l}
		\multicolumn{1}{c}{ \textbf{Chomsky-Greibach Second-Order Definite Normal Form} }\\
		\toprule
		\textbf{(Identity)} $P(x,y) \leftarrow Q(x,y):$ \\
		$target(P) $ $\wedge \; background(Q) \vee empty(Q) $ \\
		\midrule
		\textbf{(Chain)} $P(x,y) \leftarrow Q(x,z), R(z,y):$ \\
		$P \neq Q $ $\wedge \; ( target(P) \vee invented(P) ) $ $\wedge \; not(target(Q)) $ \\
		$\wedge \; not(empty(Q,R)) $ $\wedge \; invented(P,Q) \rightarrow P \neq Q $\\
		\midrule
		\textbf{(Tri-Chain)} $P(x,y) \leftarrow Q(x,z), R(z,u),	S(u,y):$ \\
		$ P \neq Q$ $\wedge \; Q \neq R$ $\wedge \; R \neq S$\\
		$\wedge \; ( target(P) \vee invented(P)$ ) $\wedge \; not(target(Q) $ \\
		$\wedge \; not(empty(Q,R,S))$ $\wedge \; invented(P,Q) \rightarrow P \neq Q $\\
	\bottomrule
	\end{tabularx}
	\caption{Chomsky-Greibach Normal Form for CFL DCGs.}
	\label{tab:full_cgnf_app}
\end{table}

\begin{table*}[!htbp]
	\centering
	\begin{tabular}{llll}
		\textbf{CNF} & \textbf{DCG} & \textbf{Definite Clauses} & \textbf{C-GNF metarules} \\
		\toprule
		$s \rightarrow \epsilon$ & $s \rightarrow empty.$    & $s(x,y) \leftarrow empty(x,y).$           & \textbf{Identity} $P(x,y) \leftarrow Q(x,y)$ \\
		$N_0 \rightarrow N_1N_2$ & $n_0 \rightarrow n_1,n_2. $ & $n_0(x,y) \leftarrow n_1(x,z), n_2(z,y).$ & \textbf{Chain} $P(x,y) \leftarrow Q(x,z), R(z,y)$ \\
		$N_i \rightarrow t$      & $n_i \rightarrow t.$      & $n_i(x,y) \leftarrow t(x,y).$             & \textbf{Identity} $P(x,y) \leftarrow Q(x,y)$ \\
					 & $empty \rightarrow [\;].$   & $empty(x,x)$ 				 & None (1st order BK) \\
					 & $t \rightarrow [t].$	     & $t([t|x],x)$ 				 & None (1st order BK) \\
		\bottomrule
	\end{tabular}
	\caption{Chomsky Normal Form mapping to DCGs and C-GNF constrained
	metarules. $N_i, n_i$: non-terminals; $t$ : terminals.}
	\label{tab:cnf_app}
\end{table*}

We present the Chomsky-Greibach Second Order Definite Normal Form (C-GNF) used
in experiments with Context-Free Grammars (CFGs) in Section 5.2 and prove that
it is, indeed, a Second Order Definite Normal Form for CFGs in Definite Clause
Grammars (DCG) notation. C-GNF is listed initially in Table \ref{tab:full_cgnf}
and listed again in Table \ref{tab:full_cgnf_app} in this Appendix for ease of
reference.

C-GNF is based on a combination of Chomsky Normal Form (CNF) and Greibach Normal
Form (GNF), two normal forms for CFGs known from the study of formal languages
and automata \cite{HopcroftUllman1969}. Every CFG can be re-written into CNF,
or GNF, or, indeed, both. GNF in particular eliminates left-recursions, and the
consequent inefficiencies when parsing CFGs with push-down automata
\cite{Greibach1965}. 

Every production in a CFG in CNF is in one of the forms shown in the first
column of Table \ref{tab:cnf}, where $N_0, N_1, N_2, N_i$ are nonterminals,
$N_i$ is a pre-terminal, $t$ is a terminal, and $\epsilon$ is the empty string.

Nonterminal symbols in the right-hand side of a CNF production cannot be the
start symbol, a constraint trivially satisfied, when converting a CFG into CNF,
by creating a new start symbol $S_0$ and expanding it to the original start
symbol $S_0 \rightarrow S$. Since this is a trivial constraint we ignore it in
C-GNF.


Additionally pre-terminal productions of the form $N_i \rightarrow t$ in CNF,
are left to be defined in the first-order background theory of an SS-ILP
learning problem\footnote{Such nonterminals can be extracted automatically from
the ground terms in training examples.}.

Otherwise, C-GNF metarule constraints enforce CNF, as listed in Table
\ref{tab:full_cgnf_app}. In the Table, a second-order variable marked as a target
(e.g. $target(P)$) is to be instantiated to the start symbol of a DCG which is
also the predicate symbol of the labelled training examples. Second-order
variables marked as background ($e.g. background(Q)$) are instantiated to
symbols in the first-order background theory, defining pre-terminals, with
terminal symbols extracted from ground terms in labelled and unlabelled
examples.

C-GNF allows second-order variables to be instantiated to invented predicate
symbols. Predicate invention is necessary for the construction of nonterminals
with exactly two symbols on the right-hand side.

GNF productions are all of the form $N_0 \rightarrow t, N_1, ... $ i.e. the
first symbol on the right-hand side of each production must be a terminal, thus
no left-recursion is possible. C-GNF metarule constraints adopt a variant of GNF
to enforce anti-left recursion assumptions while allowing for terminals to be
defined in the first-order background theory (as pre-terminals), and for
invented predicates. Thus e.g. $P \neq Q$, a constraint on \emph{Chain} and
\emph{Tri-Chain}, allows $Q$ to be instantiated to a pre-teraminal symbol, or an
invented symbol, but not the symbol of the head literal of a metarule instance.
Expanding to a pre-terminal is as incapable of producing left-recursions as
expanding to a terminal, therefore this constraint suffices to eliminate
immediate left-recursions, however it is still possible for left-recursions to
occur between invented predicates.

The constraint $invented(P,Q) \rightarrow P \neq Q$ eliminates clauses of
invented predicates that may result in "oblique" left-recursions. Consider a
pair of clauses $inv_1 \rightarrow inv_2, ..., $ and $inv_2 \rightarrow inv_1,
..., $.  These two clauses would enter an infinite left-recursion when
interpreted top-down by a standard Prolog execution strategy, even though
neither is immediately left-recursive. Such constructions are eliminated by the
above constraint.

\subsection{Completeness of C-GNF for Context-Free DCGs}
\label{Completeness of C-GNF for Context-Free DCGs}

We now show that C-GNF is a Second-Order Definite Normal Form for DCG
definitions of Context-Free Languages (CFLs).

\begin{proposition} C-GNF is a Second Order Definite Normal Form for all
	two-character, Context-Free Definite Clause Grammars.
\end{proposition}

\begin{proof}

We give a constructive proof.

The simplest way to show that a set $\mathcal{M}$ of constrained metarules is a
Normal Form for a target predicate $\Theta$ is to show that $\mathcal{M}$ can be
used to construct a maximally general logic program definition of $\Theta$.

Let $\Theta$ be a predicate $s(X,Y)$ where each of $X,Y$ is a list of
characters in $\{0,1\}$, in other words $\Theta$ represents the language of all
bit-strings. Progam $P$ in Table \ref{tab:mouth_of_god} is a maximally general
definition of $\Theta$ as a DCG. It is obvious, thus requires no further proof,
that $P$ can indeed accept all bit-strings of any length and with characters
$1$ and $0$ in any order. 

\begin{table}[t]
	\small
	\centering
	\begin{tabularx}{\columnwidth}{lll}
		\textbf{Metarule} & \textbf{Definite Clause}              & \textbf{DCG} \\
		\toprule
		\textbf{Identity} & $s(x,y) \leftarrow empty(x,y)$	  & $s \rightarrow empty$ \\
		\textbf{Chain} 	  & $s(x,y) \leftarrow one(x,z), s(z,y)$  & $s \rightarrow one$ \\ 
		\textbf{Chain}    & $s(x,y) \leftarrow zero(x,z), s(z,y)$ & $s \rightarrow zero$ \\

	  	\bottomrule
	\end{tabularx}
	\caption{Maximally-General bit-string grammar.}
	\label{tab:mouth_of_god}
\end{table}

Program $P$ in Table \ref{tab:mouth_of_god} can be constructed by substituting
the Second-Order variables in the constrained metarules \emph{Chain} and
\emph{Identity}, while satisfiying the metarule constraints imposed on
\emph{Chain} and \emph{Identity} by C-GNF, as follows.

\begin{itemize} \item To construct clause $s(x,y) \leftarrow empty(x,y)$ apply a
			substitution $\vartheta_1 = \{P/s, Q/empty\}$ to
			\emph{Identity}: $P(x,y) \leftarrow Q(x,y)\vartheta =
			s(x,y) \leftarrow empty(x,y)$.
		\item To construct clause $s(x,y) \leftarrow one(x,z), s(z,y)$
			apply a substitution $\vartheta_2 = \{P/s, Q/one, R/s\}$ to
			\emph{Chain}: $P(x,y) \leftarrow Q(x,z), R(z,y)\vartheta =
			s(x,y) \leftarrow one(x,z), s(z,y)$.
		\item To construct clause $s(x,y) \leftarrow zero(x,z), s(z,y)$
			apply $\vartheta_3 = \{P/s, Q/zero, R/s\}$ to \emph{Chain}:
			$P(x,y) \leftarrow Q(x,z), R(z,y)\vartheta = s(x,y)
			\leftarrow zero(x,z), s(z,y)$.
		\item  $\vartheta_1 = \{P/s, Q/empty\}$ satisfies the constraint
			$target(P)$ $\wedge \; background(Q)$ $\vee empty(Q)$
			imposed on \emph{Identity} substitutions by C-GNF
			because $target(s)$ is true and $empty(empty)$ is true. 
		\item $\vartheta_2 = \{P/s, Q/one, R/s\}$ satisfies the constraint
			$P \neq Q $ $\wedge \; ( target(P) \vee invented(P) ) $
			$\wedge \; not(target(Q))$ $\wedge \; not(empty(Q,R)) $
			$\wedge \; invented(P,Q)$ $\rightarrow P \neq Q $ imposed
			on \emph{Chain} by C-GNF because $s \neq one$ is true,
			and $target(s) \vee invented(s)$ is true because
			$target(s)$ is true, and $not(target(one))$ is true and
			$not(empty(one,s))$ is true, and $invented(s,one)
			\rightarrow s \neq one$ is true because
			$invented(s,one)$ is false. 
		\item $\vartheta_3 = \{P/s, Q/zero, R/s\}$ satisfies the
			aforementioned constraint imposed on \emph{Chain} by
			C-GNF because $s \neq zero$ is true, and $target(s) \vee
			invented(s)$ is true because $target(s)$ is true, and
			$not(target(zero))$ is true and $not(empty(zero,s))$ is
			true, and $invented(s,zero) \rightarrow s \neq zero$ is
			true because $invented(s,zero)$ is false. 

\end{itemize}

\end{proof}

Clauses in $P$ are instances of only two of the metarules in C-GNF, \emph{Chain}
and \emph{Identity} but this does not detract from its generality. Indeed, the
purpose of the third metarule, \emph{Tri-Chain}, is to facilitate the
construction of productions where the \emph{second} body literal is a literal of
the target predicate (e.g. $s \rightarrow one, s, one$, a production that might
appear in a grammar of palindromic bit-strings). An equivalent production could
be constructed by two instances of chain, (e.g. $s \rightarrow one, inv_1$,
$inv_1 \rightarrow s, one$) if not for the constraint $not(target(Q))$ on
\emph{Chain} used to avoid left-recursions (e.g. $inv_1 \rightarrow s, one$
violates that constraint). It is still possible to construct a more complex
program with multiple clauses of invented predicates in place of a single
instance of \emph{Tri-Chain} but at a higher computational cost. Thus, the
inclusion of \emph{Tri-Chain} improves efficiency, rather than completeness.

\subsubsection{Extending C-GNF to more than two characters}

It is easy to see that $P$ can further be extended to accept all strings of
an arbitrary number $n$ of characters: it suffices to add, for each character, a
new non-terminal instance of \emph{Chain}, and a pre-terminal in the first-order
background theory, for each new character, i.e. a total of $2n$ new productions.
Therefore C-GNF is a Second Order Normal Form for any target predicate
representing a Context-Free Language.

\section{Lindenmayer Second Order Definite Normal Form}
\label{Lindenmayer Second Order Definite Normal Form}

\begin{table*}[t]
	\centering
	\begin{tabularx}{\textwidth}{lX}
		\multicolumn{2}{c}{ \textbf{Lindenmayer Normal Form} }\\
		\toprule
		\textbf{(LS-Base)} $P(x, y, y) \leftarrow Q(x,y):$ 			& $target(P) $ $\wedge \; empty(Q) $ \\
		\textbf{(LS-Constant)} $P(x,y,z) \leftarrow Q(y,u), Q(x,v), P(v,u,z):$ 	& $target(P) $ $\wedge \; background(Q) $ $\wedge \; not(empty(Q)) $\\
		\textbf{(LS-Variable)} $P(x,y,z) \leftarrow Q(y,u), R(x,v), P(v,u,z):$ 	& $target(P) $ $\wedge \; background(Q) $ $\wedge \; not(target(R)) $ \\
		\textbf{(Chain)} $P(x,y) \leftarrow Q(x,z), R(z,y):$ 			& $invented(P) $ $\wedge \; background(Q) $ $\wedge \; not(target(R)) $ \\ 
											& $\wedge \; not(empty(Q,R))$ $\wedge \; invented(P,R) \rightarrow R \geq P $ \\
		\textbf{(Tri-Chain)} $P(x,z) \leftarrow Q(x,y), R(y,u), S(u,z):$ 	& $P \neq S $  $\wedge \; invented(P) $ $\wedge \; background(Q,R) $  \\
											& $\wedge \; not(target(S)) $ $\wedge \; not(empty(Q,R,S))$ \\
											& $\wedge \; invented(P,S) \rightarrow S \geq P $\\
											
		\bottomrule
	\end{tabularx}
	\caption{Lindenmayer Normal Form for L-System grammars as DCGs.}
	\label{tab:lnf}
\end{table*}

Lindenmayer Systems, more commonly known as L-Systems are a grammar formalism
used to describe shapes with branching and self-similar structures, like
fractals and plants \cite{ABoP1996}. L-Systems can be executed as generators for
strings of symbols interpreted as drawing commands for a Turtle graphics
interpreter, e.g. the one in the Python package
\emph{turtle}\footnote{https://docs.python.org/3/library/turtle.html}. That way,
the output of an L-system can be visualised as a shape drawn by a ``Turtle''.

Table \ref{tab:lnf} lists \emph{Lindenmayer Normal Form} (LNF) the Second Order
Normal Form used in experiments with L-System DCGs in Section 5.3. Like
L-Systems, LNF is named after Aristide Lindenmayer, who first described
L-Systems \cite{Lindenmayer1968a,Lindenmayer1968b}.

L-Systems are slightly different to phrase-structure grammars like CFGs in that
each production is applied to an input string \emph{simultaneously},
transforming the string end-to-end and completing one ``generation" at a time
\cite{ABoP1996}. The string produced in each generation is passed to the grammar
again in the next generation and again processed by all productions
simultaneously.

\subsection{Constructing an L-System DCG}
\label{Defining an L-System DCG}

\begin{table}[t]
	\small
	\setlength{\tabcolsep}{1.6mm}
	\centering
	\begin{tabularx}{\columnwidth}{ll}
		\multicolumn{2}{c}{ \textbf{Dragon Curve L-System DCG} }\\
		\toprule
		\textbf{(1a) L-System notation}   & \textbf{(1b) DCG notation} \\ 
		\toprule
		Variables: $f,g$             & $s([+|Ss]) \rightarrow plus, s(Ss)$ \\ 
		Constants: $+,-$             & $s([-|Ss]) \rightarrow minus, s(Ss)$ \\
		$\delta$: 90$\degree$        & $s([f,+,g|Ss]) \rightarrow f, s(Ss)$ \\
		$f \rightarrow f+g$          & $s([f,-,g|Ss]) \rightarrow g, s(Ss)$ \\
		$g \rightarrow g-f$          & $s([\;]) \rightarrow [\;]$ \\ 
	\end{tabularx}
	\begin{tabularx}{\columnwidth}{l}
		\midrule
		\textbf{(2) Definite Clauses} \\
		\toprule
		$s([+|x], y, z) \leftarrow plus(y, u),  s(x, u, z).$  \\
		$s([-|x], y, z) \leftarrow minus(y, u), s(x, u, z).$   \\
		$s([f, +, g|x], y, z) \leftarrow f(y, u), s(x, u, z).$ \\
		$s([f, -, g|x], y, z) \leftarrow g(y, u), s(x, u, z).$ \\
		$s([], X, Y) \leftarrow X=Y.$ \\
		\midrule
	\end{tabularx}
	\begin{tabularx}{\columnwidth}{l}
		\textbf{(3) Definite Clauses (flattened, with invented
		symbols)} \\
		\toprule
		$s(x,y,z) \leftarrow plus(y,u),plus(x,v),s(v,u,z)$ \\
		$s(x,y,z) \leftarrow minus(y,u),minus(x,v),s(v,u,z)$ \\
		$s(x,y,z) \leftarrow f(y,u),inv\_1\_29(x,v),s(v,u,z)$ \\
		$inv\_1\_29(x,y) \leftarrow f(x,z),plus(z,u),g(u,y)$ \\
		$s(x,y,z) \leftarrow g(y,u),inv\_1\_36(x,v),s(v,u,z)$ \\
		$inv\_1\_36(x,y) \leftarrow f(x,z),minus(z,u),g(u,y)$ \\
		$s(x,y,y) \leftarrow empty(x,y)$ \\
		\midrule
	\end{tabularx}
	\begin{tabularx}{\columnwidth}{X}
		\textbf{(4) Definite Clauses (flattened,unfolded to remove
		invented symbols)} \\
		\midrule
		$s(x,y,z) \leftarrow plus(y,u),plus(x,v),s(v,u,z)$ \\
		$s(x,y,z) \leftarrow minus(y,u),minus(x,v),s(v,u,z)$ \\
		$s(x,y,z) \leftarrow f(y,u)$, $f(x,v)$, $plus(v,w)$, $g(w,g)$, $s(g,u,z)$ \\
		$s(x,y,z) \leftarrow g(y,u)$, $f(x,v)$, $minus(v,w)$, $g(w,g)$, $s(g,u,z)$ \\
		$s(x,y,y) \leftarrow empty(x,y)$ \\
		\midrule
	\end{tabularx}
	\begin{tabularx}{\columnwidth}{lll}
		\multicolumn{2}{c}{ \textbf{(5) First-order background theory} }\\
		\midrule
		\textbf{Definite clauses} & \textbf{DCG} & \textbf{Drawing command}  \\
		$plus([+|x],x)$  & $plus \rightarrow [+]$ & ``Turn right 90$\degree$'' \\
		$minus([-|x],x)$ & $minus \rightarrow [-]$ & ``Turn left 90$\degree$'' \\
		$f([f|x],x)$     & $f \rightarrow [f]$     & ``Move forward (1)'' \\
		$g([g|x],x)$     & $g \rightarrow [g]$     & ``Move forward (2)'' \\
		\bottomrule
	\end{tabularx}
	\caption{Dragon Curve L-System used in Section 5. Symbols $+,-$ are not
	mathematical functions but drawing commands for a Turtle graphis
	interpreter.}
	\label{tab:dragon_curve}
\end{table}

Similar to phrase-structure grammars L-Systems are rewrite systems: their
productions determine how characters in a string are replaced with other
characters, or strings. L-System strings consist of two kinds of characters,
constants and variables, analogous to terminals and non-terminals,
respectively. Constants are only replaced by themselves while variables are
replaced by arbitrary strings of constants and variables, depending on the
L-System.

When considering the possible structure of an L-System DCG, the requirement
that all productions must be applied to a string simultaneously suggests a DCG
where every production is recursive, save one that expands to the empty string;
that way, a string must be consumed in full in order to be accepted by a
grammar. This intuition informs the recursive structure of metarules in LNF.

The distinction of symbols in constants (terminals) and variables (nonterminals)
suggests at least two metarules are needed: one for constants and one for
variables. Those are the metarules \emph{LS-Constant} and \emph{LS-Variable},
respectively, listed in Table \ref{tab:lnf}. In both these metarules, the first
body literal $Q(y,u)$ is a literal of a DCG pre-terminal (defined in a
first-order background theory) that corresponds to the symbol consumed in the
input, i.e.  either a constant or a variable. If the input symbol is a constant
the second body literal $Q(x,v)$ in \emph{LS-Constant} is a literal of the same
pre-terminal as $Q(y,u)$, thus instances of the \emph{LS-Constant} metarule can
only represent productions that replace a constant with itself. If the input
symbol is a variable, the second body literal, $R(x,v)$ in \emph{LS-variable},
can be a literal of the same or a different predicate than in $Q(y,u)$. In
particular, $R(x,v)$ can be a literal of an invented predicate (but not the
target predicate, i.e. $s/3$ itself; LNF only allows tail-recursion). If
$R(x,v)$ is a literal of an invented predicate, clauses of this predicate
\emph{must} be instances of either \emph{Chain} or \emph{Tri-Chain}. 

The combination of \emph{Chain} and \emph{Tri-Chain} allows the construction of
a sequence of clauses of invented predicates capable of pushing into the output
a string of any length. In particular, instances of \emph{Chain} can add at most
two symbols to the output and terminate the recursion, or add one symbol and
continue the recursion, while instances of \emph{Tri-Chain} can add at most
three symbols to the output and terminate the recursion, or add two symbols and
continue the recursion. In both cases, recursion is continued with a
tail-recursive call from the final body literal of an instance of \emph{Chain}
or \emph{Tri-Chain}, this literal having a new invented predicate symbol that
must be higher in alphabetic order than the invented predicate symbol in the
head of the same \emph{Chain} or \emph{Tri-Chain} instance (constraints
$invented(P,R) \rightarrow R \geq P$, in \emph{Chain}, and $invented(P,S)
\rightarrow S \geq P$ in \emph{Tri-Chain}). This strong constraint eliminates
redundant constructions of invented predicates that differ syntactically only in
the names, and the order, of their invented symbols.

\subsection{A Dragon System DCG}
\label{A Dragon System DCG}

The effect of imposing LNF on hypotheses constructed by Poker (Procedure
\textsc{Generalise} in Algorithm 1) is illustrated in Table
\ref{tab:dragon_curve}. In the Table, the top row lists the Dragon Curve
L-System in the notation common in L-System literature, in the column marked
(1a), and manually encoded as a DCG in the column marked (1b); this DCG was used
to generate examples for experiments in the Experiments Section. The second row
of the Table, marked (2), lists the plain definite clause notation (without DCG
``syntactic sugar") to clarify the pattern of variable sharing between literals.

The third row in Table \ref{tab:dragon_curve}, marked (3), lists a correct
hypothesis learned by Poker\footnote{It should be noted that Poker did not
learn this hypothesis all in one go, but piecemeal. Two sub-sets of that
hypothesis, each with 4 clauses, were constructed separately, as in Line 2 in
Algorithm 1, then combined by taking their union as in Line 19 of the
Algorithm. The ability to learn programs in chunks allows Poker to learn larger
programs with many invented predicates efficiently.}. This hypothesis is
(success-set) equivalent to the DCG in column (1b) of the top row of the Table, 
and therefore also the set of definite clauses in row (2). Syntactically the
program in row (3) differs to that in row (2) in that the list of constants in the
first argument of each head literal in the program in row (2) is replaced, in
row (3), with body literals of the corresponding pre-terminals with suitable
variable bindings. Thus, the program in row (3) is, in effect, a
\emph{flattened} version of the program in row (2) of the Table, in the sense of
flattening described in \cite{Rouveirol1994}\footnote{This kind of flattening is
used in the ILP literature to remove function symbols, such as the Prolog
list-constructor symbol $|$, from literals of learned hypotheses, so that the
language of hypotheses is a function-free First Order language. The goal of this
transformation is improved learning efficiency. In MIL a function-free
hypothesis language is forced by the nature of Metarules which have no variables
quantified over non-constant function symbols.}. This flattened program is listed
with invented predicate symbols as constructed by Poker, to illustrate the
use of predicate invention when learning an L-System DCG. In particular,
predicate invention makes it possible to use metarules with a constant number of
literals to learn grammars with a variable number of literals, albeit in
``folded`` form. Indeed, the program in Row (3) is a folded form of the program
in row (4).

The fourth row in Table \ref{tab:dragon_curve}, marked (4), lists the
hypothesis learned by Poker, as in row (3), but this time unfolded
automatically to remove invented predicates. In this form it is easier to
observe the assembly of output strings by the sequence of invented predicate
clauses in the previous row. Preterminals, representing constants used in the
Dragon Curve L-System, are listed in the bottom-most row of the Table, along
with their Turtle language interpretation as drawing commands.

\subsection{Completeness of LNF for L-Systems}
\label{Completeness of LNF for L-Systems}

\begin{table}
	\small
	\centering
	\setlength{\tabcolsep}{1.6mm}
	\begin{tabular}{lll}
		\multicolumn{3}{c}{ \textbf{Maximally general L-system Grammar}	}\\
		\toprule
		\# & \textbf{DCG}                 & \textbf{Definite Clauses} \\
		\midrule
		1  & $s \rightarrow c_1, c_1, s$  & $s(x,y,z) \leftarrow c_1(y,u), c_1(x,v), s(v,u,z)$ \\
		2  & $s \rightarrow c_2, c_2, s$  & $s(x,y,z) \leftarrow c_2(y,u), c_2(x,v), s(v,u,z)$ \\
		3  & $s \rightarrow v_1, c_1, s$  & $s(x,y,z) \leftarrow v_1(y,u), c_1(x,v), s(v,u,z)$ \\ 
		4  & $s \rightarrow v_1, c_2, s$  & $s(x,y,z) \leftarrow v_1(y,u), c_2(x,v), s(v,u,z)$ \\
		5  & $s \rightarrow v_1, v_1, s$  & $s(x,y,z) \leftarrow v_1(y,u), v_1(x,v), s(v,u,z)$ \\
		6  & $s \rightarrow v_1, v_2, s$  & $s(x,y,z) \leftarrow v_1(y,u), v_2(x,v), s(v,u,z)$ \\
		7  & $s \rightarrow v_2, v_1, s$  & $s(x,y,z) \leftarrow v_2(y,u), v_1(x,v), s(v,u,z)$ \\
		8  & $s \rightarrow v_2, v_2, s$  & $s(x,y,z) \leftarrow v_2(y,u), v_2(x,v), s(v,u,z)$ \\
		9  & $s \rightarrow v_2, c_1, s$  & $s(x,y,z) \leftarrow v_2(y,u), c_1(x,v), s(v,u,z)$ \\
		10 & $s \rightarrow v_2, c_2, s$  & $s(x,y,z) \leftarrow v_2(y,u), c_2(x,v), s(v,u,z)$ \\
		11 & $s \rightarrow empty$        & $s(x,y,y) \leftarrow empty(x,y)$ \\
	\end{tabular}
	\begin{tabularx}{\columnwidth}{lll}
		\midrule
		\multicolumn{3}{c}{ {\hspace{3em}} \textbf{First-Order Background Theory (pre-terminals)} } \\
		\midrule
		& \textbf{DCG} & \textbf{Definite Clauses} \\
		\midrule
		Constants: & $c_1 \rightarrow [c_1]$  & $c_1([c_1|x],x)$. \\ 
			   & $c_2 \rightarrow [c_2]$  & $c_2([c_2|x],x)$. \\ 
		Variables: & $v_1 \rightarrow [v_1]$  & $v_1([v_1|x],x)$. \\ 
			   & $v_2 \rightarrow [v_2]$  & $v_2([v_2|x],x)$. \\
		\bottomrule
	\end{tabularx}
	\caption{Maximally general grammar for L-Systems with two constant and
	two variable symbols constructed according to LNF.}
	\label{tab:mouth_of_Lindenmayer}
\end{table}

We now give a proof that LNF is a Second Order Definite Normal Form for
L-Systems. As with C-GNF we give a proof by construction, of a maximally general
L-System DCG listed in Table \ref{tab:mouth_of_Lindenmayer}. Our DCG assumes an
L-System with only two constants, and two variable symbols, $c_1,c_2$ and
$v_1,v_2$, respectively. These can be trivially interpreted as any desired
characters, such as $+,-,f,g$ in the Dragon Curve L-System in Table
\ref{tab:dragon_curve}. 

\begin{proposition} LNF is a Second Order Definite Normal Form for all Definite
	Clause Grammars of L-System languages with two variables and two
	constants.
\end{proposition}

As for C-GNF we give a constructive proof.

\begin{proof}

The grammar in Table \ref{tab:mouth_of_Lindenmayer} includes one instance of the
\emph{LS-Constant} metarule for each of the two constants, $c_1,c_2$, consuming
from the input, and then pushing into the output, the corresponding constant. It
includes four instances of the \emph{LS-Variable} metarule for each of the two
variables, $v_1,v_2$, each of which consumes the corresponding variable, and
outputs one other symbol, each of the two constants of variables. An instance of
\emph{LS-Base} completes the DCG. 

It should be obvious, and so should need no further poof, that the DCG in Table
\ref{tab:mouth_of_Lindenmayer} accepts all strings of the symbols in
$\{c_1,c_2,v_1,v_2\}*$ of arbitrary length and arbitrary order, i.e. this
grammar is a maximally general grammar for L-Systems with two constants and two
variables (suitably substituted for $c_1,c_2$ and $v_1,v_2$, respectively). 

The DCG in Table \ref{tab:mouth_of_Lindenmayer} can be constructed by
substituting the Second-Order variables in the constrained metarules
\emph{LS-Base}, \emph{LS-Constant}, \emph{LS-Variable}, \emph{Chain} and
\emph{Tri-Chain}, while satisfying the metarule constraints imposed on the
aforementioned metarules by LNF, below. In the following, for brevity, we
treat the construction of similar clauses together by means of variable indices
in place of the indices 1, 2 in symbols $c_1, c_2, v_1, v_2$.


\begin{itemize} \item To construct clauses 1, 2  in Table
			\ref{tab:mouth_of_Lindenmayer} (i.e. clauses
			corresponding to productions consuming constant symbols)
			apply a substitution $\vartheta_i = \{P/s,Q/c_i\}$ to
			\emph{LS-Constant}, $\forall i \in [1,2]$. $P(x,y,z)$
			$\leftarrow$ $Q(y,u)$, $Q(x,v)$, $P(v,u,z)\vartheta_i$ =
			$s(x,y,z)$ $\leftarrow$ $c_i(y,u)$, $c_i(x,v)$,
			$s(v,u,z)$, where $i$ is 1 for clause 1, and 2 for
			clause 2.
		\item To construct clauses 3-4 and 9-10 in Table
			\ref{tab:mouth_of_Lindenmayer} (i.e. clauses
			corresponding to productions consuming variable symbols
			and replacing them with constant symbols) apply a
			substitution $\varphi_j^k = \{P/s$,$Q/v_j$,$R/c_k\}$
			to \emph{LS-Variable}, $\forall j,k \in [1,2]$.
			$P(x,y,z)$ $\leftarrow$ $Q(y,u)$, $R(x,v)$,
			$P(v,u,z)\varphi_j^k$ = $s(x,y,z)$ $\leftarrow$
			$v_j(y,u)$, $c_k(x,v)$, $s(v,u,z)$.
		\item To construct clauses 5-8 in Table
			\ref{tab:mouth_of_Lindenmayer} (i.e. clauses
			corresponding to productions consuming variable symbols
			and replacing them with variable symbols) apply a
			substitution $\sigma_j^k = \{P/s$,$Q/v_j$,$R/v_k\}$
			to \emph{LS-Variable}, $\forall$ $j,k$ $\in [1,2]$.
			$P(x,y,z)$ $\leftarrow$ $Q(y,u)$, $R(x,v)$,
			$P(v,u,z)\sigma_j^k$ = $s(x,y,z)$ $\leftarrow$
			$v_j(y,u)$, $v_k(x,v)$, $s(v,u,z)$.
		\item To construct clause 11 in Table
			\ref{tab:mouth_of_Lindenmayer} apply a substitution
			$\omega = \{P/s,Q/empty\}$ to \emph{LS-Base}. $P(x, y,
			y) \leftarrow Q(x,y)\omega = s(x,y,y) \leftarrow
			empty(x,y)$.
		\item Substitution $\vartheta_i = \{P/s,Q/c_i\}$ satisfies the
			constraint $target(P)$ $\wedge$ $background(Q)$ $\wedge$
			$not(empty(Q))$ imposed on \emph{LS-Constant} by LNF
			because $target(s)$ is true and $background(c_i)$
			$\wedge$ $not(empty(c_i))$ is true for $c_i$ $\in$
			$\{c_1,c_2\}$.
		\item Substitution $\varphi_j^k = \{P/s$,$Q/v_j$,$R/c_k\}$
			satisfies the constraint $target(P)$ $\wedge$
			$background(Q)$ $\wedge$ $not(target(R))$ imposed on
			\emph{LS-Variable} by LNF because $target(s)$ is true
			and $background(v_j)$ is true for $v_j$ $\in$ $\{v_1,
			v_2\}$ and $not(target(c_k))$ is true for $c_k$ $\in$
			$\{c_1,c_2\}$.
		\item Substitution $\sigma_j^k = \{P/s$,$Q/v_j$,$R/v_k\}$
			satisfies the constraint $target(P)$ $\wedge$
			$background(Q)$ $\wedge$ $not(target(R))$ imposed on
			\emph{LS-Variable} by LNF because $target(s)$ is true
			and $background(v_j)$ is true for $v_j$ $\in$
			$\{v_1,v_2\}$ and $not(target(v_k))$ is true for $v_k$
			$\in$ $\{v_1,v_2\}$.
		\item Substitution $\omega = \{P/s,Q/empty\}$ satisfies the
			constraint $target(P)$ $\wedge$ $empty(Q)$ imposed on
			\emph{LS-Base} by LNF because $target(s)$ is true and
			$empty(empty)$ is true.
\end{itemize}

\end{proof}

As with C-GNF, the non-inclusion of instances of \emph{Chain} and
\emph{Tri-Chain} in the grammar listed in Table \ref{tab:mouth_of_Lindenmayer}
does not detract from its generality. The purpose of \emph{Chain} and
\emph{Tri-Chain} is to allow learning DCG productions with variable numbers of
body literals in ``folded" form, as discussed above.

\subsubsection{Extending LNF to L-System languages with more than two variables
and constants}

We have shown that LNF is a Second Order Definite Normal Form for DCGs of
L-System languages with two constants and two variables. The same result can be
extended to L-Systems with any number of constants or variables by adding, or
removing, one instance of \emph{LS-Constant} for each added or removed constant
symbol, and $k$ instances of \emph{LS-Variable} for each added or removed
variable symbol, where $k$ is the cardinality of the set of symbols in the
represented L-System.

\section{Experimental setup}
\label{Experimental setup}

All experiments reported in the Experiments section were carried out on a server
with two AMD EPYC 7551 CPUs with a clock speed of 2.0GHz, 32 cores, and 256GB of
RAM, running Linux Fedora 41.

The values of Poker and Louise hyperparameters used in the experiments reported
in the Experiments section can be found in the joint Code Appendix, where
experiment scripts are defined as Prolog predicates that set hyperparameter
values and generate training and testing exampes.

Hyperparameters were selected in preliminary experiments carried out on a GPD
Win Max 2 2024 mini laptop with an AMD Ryzen 7 8840U processor with 16 cores
clocked at 3.3 GHz and with 64 GB of RAM, running Linux Fedora 39. These
preliminary experiments were executed over 1 to 10 randomly sampled sets of
training and testing examples in each experiment (compared with 100 in reported
experiments).

\end{document}